\documentclass[runningheads]{ola}

\usepackage{epsfig}
\usepackage{bm}
\usepackage{amsfonts,amsmath,amssymb}
\usepackage{xcolor}
\usepackage[numbers,sort&compress]{natbib}
\usepackage{caption}
\usepackage{subcaption}

\newcommand{\R}{\mathbb{R}}

\newcommand{\N}{\mathbb{N}}

\newcommand{\lo}{\longrightarrow}

\begin{document}
\pagestyle{headings}

\mainmatter

\title{Polynomial-Model-Based Optimization for Blackbox Objectives }

\titlerunning{Polynomial-Model-Based Optimization}

\author{Janina Schreiber \and Damar Wicaksono \and Michael Hecht}

\authorrunning{Schreiber et al.}

\institute{Center for Advanced Systems Understanding (CASUS), Görlitz, Germany\\
Helmholtz-Zentrum Dresden-Rossendorf (HZDR), Dresden, Germany\\
{\email{j.schreiber@hzdr.de}, \email{d.wicaksono@hzdr.de}, \email{m.hecht@hzdr.de}}
}
\maketitle

\begin{abstract}
    For a wide range of applications the structure of systems like Neural Networks or complex simulations, is unknown and approximation is costly or even impossible. Black-box optimization seeks to find optimal (hyper-) parameters for these systems such that a pre-defined objective function is minimized. Polynomial-Model-Based Optimization (PMBO) is a novel blackbox optimizer that finds the minimum by fitting a polynomial surrogate to the objective function.

    Motivated by Bayesian optimization the model is iteratively updated according to the acquisition function Expected Improvement, thus balancing the exploitation and exploration rate and providing an uncertainty estimate of the model. PMBO is benchmarked against other state-of-the-art algorithms for a given set of artificial, analytical functions. PMBO competes successfully with those algorithms and even outperforms all of them in some cases. As the results suggest, we believe PMBO is the pivotal choice for solving blackbox optimization tasks occurring in a wide range of disciplines.\footnote{This work was presented as a peer reviewed contribution at {\em OLA'2023 International Conference on Optimization and Learning, Malaga, Spain, 2023}
 }
\end{abstract}

\section{Introduction}

Optimization tasks arise in many fields of applications, ranging from analyzing and adjusting simulations of complex systems across all disciplines to inferring optimal neural network architectures \cite{Jlassi2021, Ozik2018, Ozik2019}.
Typically, the underlying objective function of the optimization problems is unknown and very costly to sample. This is why
\emph{blackbox optimization methods} are designed to find an optimum without any a priori knowledge of the underlying function by sampling the objective function as sparsely as possible.

We contribute to this omnipresent and challenging task in scientific computing by introducing a novel blackbox optimizer, which we term \emph{Polynomial-Model-Based Optimization} (PMBO).
Motivated by Bayesian optimization \cite{Jones1998}, the generated optimization model is iteratively updated according to an acquisition function that balances the exploitation and exploration rate. The location of the desired optimum is predicted by efficiently computing the optimum of the (analytic, polynomial) optimization model.

We evaluate the performance of PMBO by benchmarking it against established state-of-the-art algorithms for commonly used analytically given objective functions up to dimension $m=6$. Loosely speaking, the \emph{No Free Lunch Theorem} \cite{wolpert1997no} states that for any two optimizers $A$, $B$ there are as ``many'' objective functions $f_A$ where $A$ outperforms $B$ as there are functions $f_B$ for which the converse is true. Thus, a universally best-performing optimizer cannot even be expected to exist in practice.

The results presented herein suggest, however, that in the problem range of the addressed instances, PMBO performs superior to \emph{Random Search} \cite{James2012}, \emph{Quasi-Random Search}, \emph{Bayesian Optimization} \cite{Snoek2012}, \emph{Particle Swarm Optimization} (PSO) \cite{Kennedy2017}  and comparable to \emph{Covariance Matrix Adaptation - Evolution Strategy (CMA-ES)} \cite{Hansen2003}.

We continue by shortly sketching  PMBO's methodology.
\begin{figure}[h]
\vspace{-5pt}
    \centering
    \subfloat \centering a)
    {{\includegraphics[width=6.25cm]{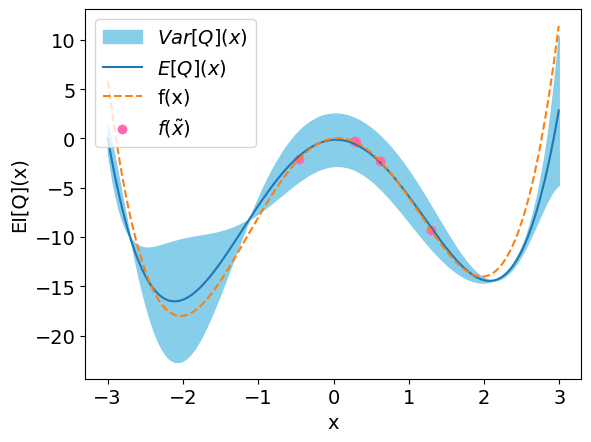} }}%
    \qquad
    \subfloat \centering b)
    {{\includegraphics[width=6.25cm]{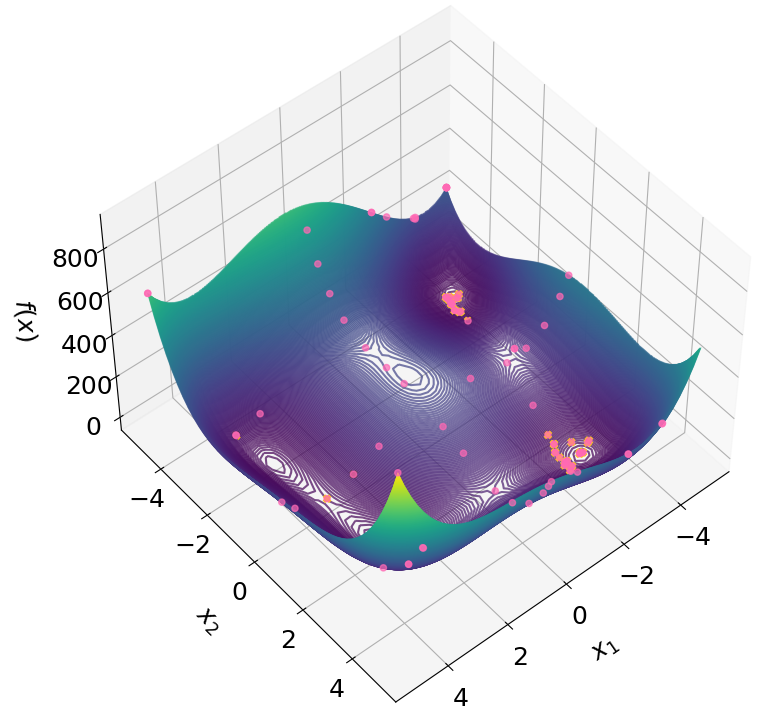} }}%
    \caption{a) Illustration of the Expected Improvement acquisition function. The ground truth function $f$ is plotted as the orange dotted line. A polynomial $Q$ is fitted through the sample points (in pink). The expected value $E[Q] =Q$ is plotted as the blue line with a band given by the uncertainty $\mathop{Var}[Q]$. b) shows the sampling strategy of PMBO for the 2D Himmelblau function \cite{Jamil2013}.}
    \label{fig:exJamil2013mple}%
\end{figure}

\section{Methodology of PMBO}

The development of PMBO was motivated by the classic \emph{Bayesian optimization} \cite{Snoek2012}.
However, instead of fitting a stochastic process, a polynomial $Q_f$ serves as the surrogate model of the objective function $f$, predicting the optimum of $f$ by localizing the optimum of $Q_f$.
The polynomial $Q_f$ is updated according to an acquisition function following an expected improvement strategy.
The update process is repeated until a convergence criterion is met or a maximum number of iterations is reached.

PMBO is implemented in Python based on the open source library {\sc minterpy} \cite{minterpy} and will become available at https://github.com/minterpy-project.

\subsection{Polynomial surrogates}

Classic $1$-dimensional (1D) polynomial interpolation goes back to Newton, Lagrange, and others \cite{Meijering2002}. Its generalization to regression tasks was
mainly proposed and developed by Gau\ss, Markov, and Gergonne \cite{gergonne1974application,stigler1974gergonne}
and is omnipresent in mathematics and computing today. Here, we use recent extensions of these classics to multi-dimensions by resisting the curse of dimensionality \cite{PIP1,PIP2,MIP,IEEE,REG}, with numerically stable and efficient implementations given in \cite{minterpy}.

Formally, the objective function $f: \Omega =[-1,1]^m\lo \R$, $m \in \N$ is approximated by a polynomial surrogate $Q(x) \in \Pi_A=\mathrm{span}\{x^\alpha= x_1^{\alpha_1}\cdots x_m^{\alpha_m}\}$ in Lagrange form
\begin{equation}
    \label{eqn:lagr}
    f(x) \approx Q(x) = \sum_{\alpha \in A} c_{\alpha}L_{\alpha}(x)\,, \quad L_{\alpha}(p_\beta) = \delta_{\alpha,\beta}\,,
\end{equation}
where the Lagrange polynomials are uniquely determined with respect to a properly chosen set of unisolvent nodes $P_A=\{p_\beta: \beta \in A\}$. The multi-index set $A\subseteq \N^m$ generalizes the concept of the polynomial degree to multi-dimensions and enables identify sparse polynomial models approximating the function $f$.
The coefficients are either given by $c_\alpha = f(p_\alpha)$ in case of Lagrange-interpolation or are derived by Newton-Lagrange regression
\cite{PIP1,PIP2,MIP,IEEE,REG}.



\subsubsection{Polynomial update}
PMBO initializes the polynomial surrogate by choosing an initial sampling seed of the objective function $f$. However, being a blackbox function, the initialization cannot be guaranteed to be properly chosen in general.
The ability to iteratively update the polynomial surrogate is therefore indispensable.

Given a polynomial surrogate as in Eq.~(\ref{eqn:lagr}), the update procedure is given by increasing the complexity of the polynomial surrogate by enlarging the multi-index set $A \mapsto A\cup\{\alpha^*\}$ due to specific choices $\alpha^* \in \N^m$. While the multi-indices $\alpha\in A$ and the unisolvent nodes $p_\alpha \in P_A$ are in one-to-one correspondence the decision, of which potential update to choose, is mainly driven by the optimum of
the acquisition function, $\alpha^* = \mathrm{argmax}_{\beta}EI(p_\beta)$.



\subsection{The acquisition function}
Motivated by Bayesian optimization the \emph{Expected Improvement} acquisition function is defined as
\begin{equation}
    EI[Q](x) = E[Q](x) - \gamma \mathop{Var}[Q](x)\,, \quad \gamma \in [0,1]\,.
\end{equation}
The expected value $E[Q](x)=Q(x)$ at a given argument $x\in \Omega=[-1,1]^m$ coincides with the value of the polynomial surrogate.
The variance $\mathop{Var}[Q](x)$ represents the uncertainty of the surrogate at a given point,
 which we model by using the bootstrapping technique \cite{Efron1979}.
The parameter  $\gamma$ controls the impact of the model uncertainty and balances the amount of exploration and exploitation, see Fig.~\ref{fig:exJamil2013mple}~a). The resulting PMBO sampling for the 2D Himmelblau function, capturing all optima,  is given in Fig.~\ref{fig:exJamil2013mple}~b).

\section{Empirical results}
For evaluating the performance of PMBO, we compare it to the aforementioned state-of-the-art alternatives for two analytic benchmark functions: The \emph{Hartmann Function} in dimension $m=3$ and the \emph{Rosenbrock function} in dimension $m=6$ \cite{Jamil2013}. The maximum number of iterations was set to $300$ for both problems and each algorithm was repeated $5$ times.
PMBO computes the initial polynomial surrogate $Q_f$ by sampling the objective function on a $50$-point seed. Different initial sampling strategies are used, such as \emph{Random sampling}, \emph{Chebyshew nodes}, \emph{Sobol sequences}, and the first $50$ \emph{CMA-ES} samples, which we denote by \emph{PMBO-Random}, \emph{PMBO-Chebyshew}, etc.
\begin{figure}[ht]
\vspace{-5pt}
    \centering
    \subfloat \centering a)
    {{\includegraphics[width=7.5cm]{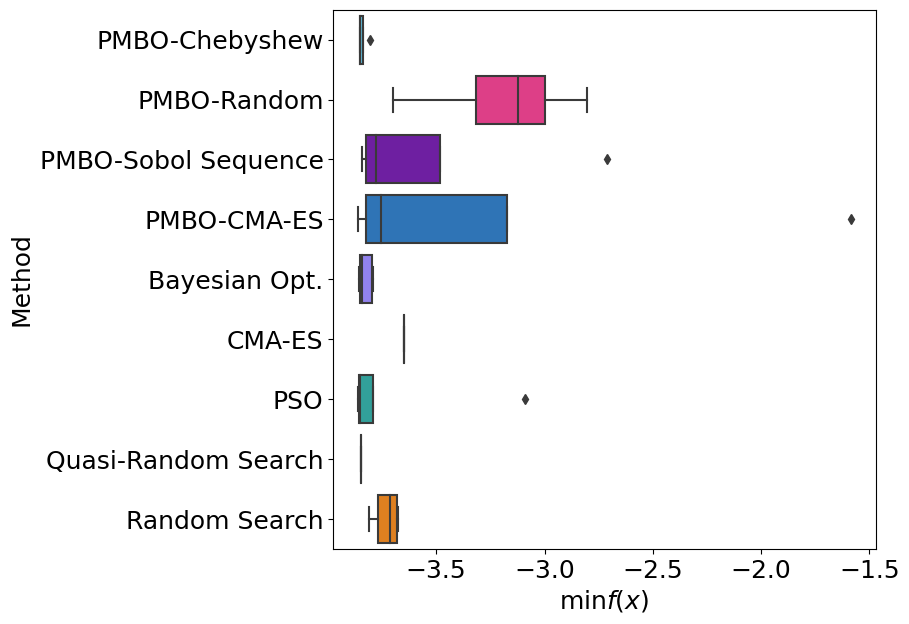} }}%
    \qquad
    \subfloat \centering b)
    {{\includegraphics[width=5.5cm]{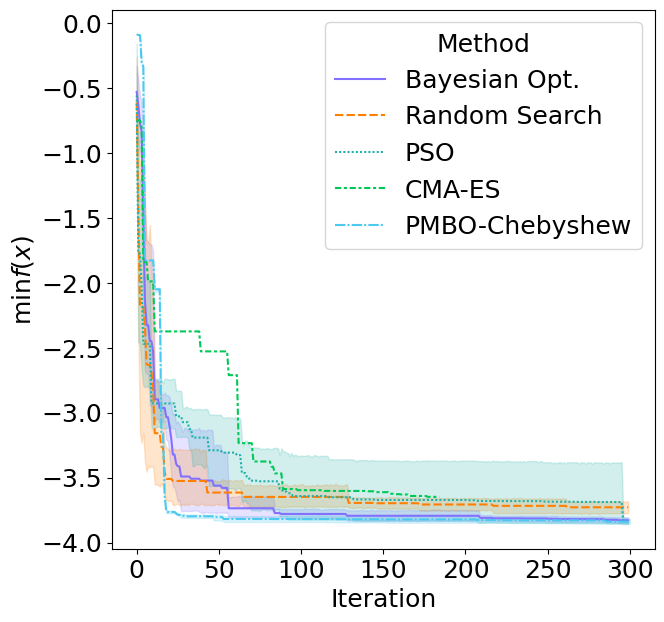} }}%
    \vspace{-5pt}
    \caption{Hartmann function in dimension $m=3$.}
    \label{fig:hartmann}%
\end{figure}

Figure \ref{fig:hartmann} reports the results for the Hartmann function in dimension $m=3$. While PMBO-Chebyshew outperforms all other methods, PMBO's performance is sensitive to the choice of the initial samples. Interestingly, PMBO-CMA-ES improves CMA-ES on average.

\begin{figure}[h!]
\vspace{-5pt}
    \centering
    \subfloat \centering a)
    {{\includegraphics[width=7.25cm]{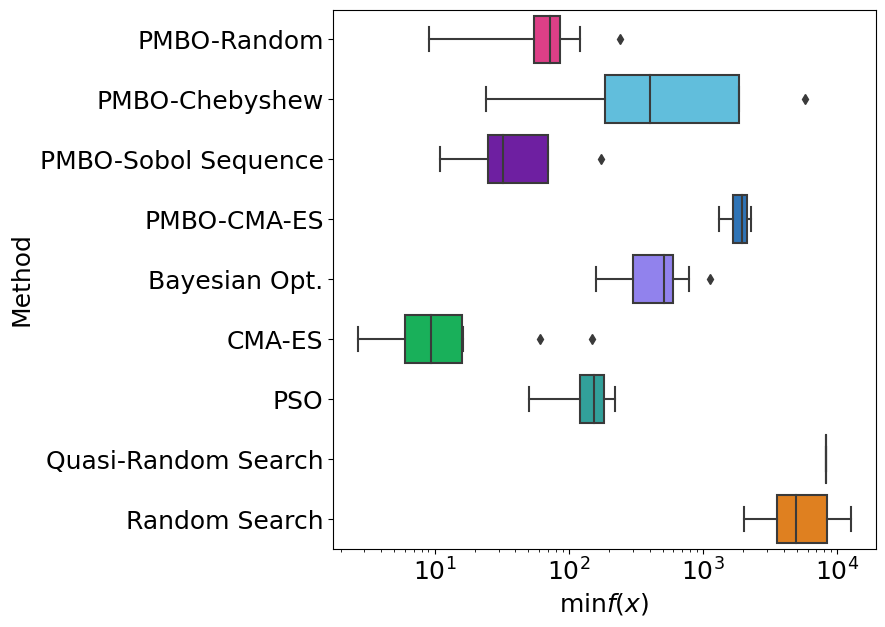} }}%
    \qquad
    \subfloat \centering b)
    {{\includegraphics[width=5.75cm]{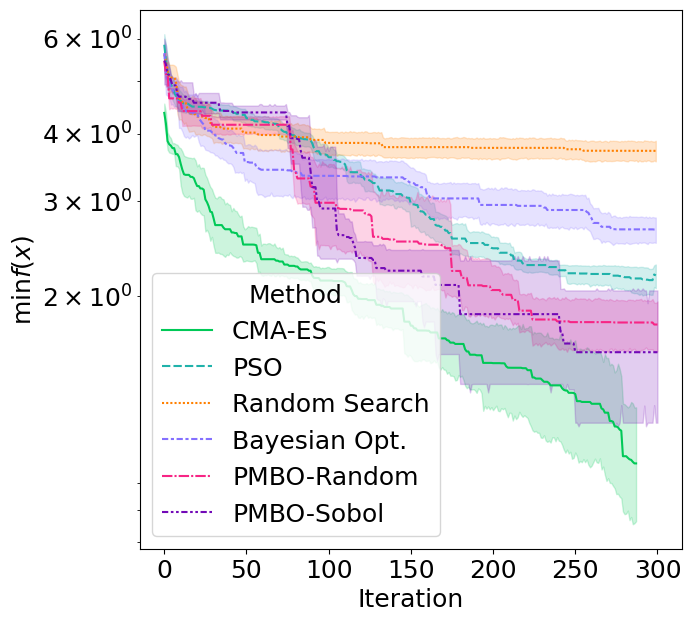} }}%
    \vspace{-5pt}
    \caption{Rosenbrock function in dimension $m=6$.}
    \label{fig:rosenbrock}%
\end{figure}
In Fig.~\ref{fig:rosenbrock} the results for the 6D Rosenbrock function are plotted. While
CMA-ES outperforms all other methods, PMBO-Random and PMBO-Sobol are 2nd best choices.
Though being a non-exhaustive evaluation, the presented results make us believe that a vast class of blackbox optimization tasks will benefit from the novel PMBO, as further discussed below.

\section{Outlook}

As part of our ongoing research, we exploit the specific nature of PMBO, enabling us to theoretically identify mild constraints on the objective function that enable optimal PMBO initialization and yield criteria for PMBO's applicability in practice.

As a real-world showcase, we currently address the hyper-parameter-tuning-task for \emph{U-Net}, \cite{Ronneberger2015}, being a central machine learning-based approach for solving image classification and segmentation tasks occurring in systems biology and bio-medicine \cite{Zhao2022}. A deeper theoretical investigation and extended empirical demonstrations is in progress, involving the uncertainty quantification test-suit \cite{DamarUQ}.

\section*{Acknowledgments} 
This work was partially funded by the Center of Advanced Systems Understanding (CASUS), financed by Germany's Federal Ministry of Education and Research (BMBF) and by the Saxon Ministry for Science, Culture and Tourism (SMWK) with tax funds on the basis of the budget approved by the Saxon State Parliament.

\bibliographystyle{plain}
\bibliography{Ref.bib}

\end{document}